\documentclass[lettersize,journal]{IEEEtran}
\usepackage{amsmath,amsfonts}
\usepackage{algorithmic}
\usepackage{algorithm}
\usepackage{array}
\usepackage[caption=false,font=normalsize,labelfont=sf,textfont=sf]{subfig}
\usepackage{textcomp}
\usepackage{stfloats}
\usepackage{url}
\usepackage{verbatim}
\usepackage{graphicx}
\usepackage{amssymb}
\usepackage{algorithmic}
\usepackage{xcolor}
\usepackage{bm}
\usepackage{latexsym}
\usepackage{multirow}
\usepackage{threeparttable}
\usepackage{amsthm}

\begin{document}

\title{Federated Learning with Hypergradient-based\\ Online Update of Aggregation Weights}

\author{Ayano Nakai-Kasai,~\IEEEmembership{Member,~IEEE,} 
        \thanks{A. Nakai-Kasai and T. Wadayama are with Nagoya Institute of Technology, Gokiso-cho, Showa-ku, Nagoya, Aichi 466-8555, Japan,}
      % Tatsuki Tokumura,~\IEEEmembership{Non-Member,} 
      and Tadashi Wadayama,~\IEEEmembership{Member,~IEEE} 
    %   \thanks{T. Wadayama is with Nagoya Institute of Technology, Gokiso, Nagoya, Aichi 466-8555, Japan,}
        % <-this % stops a space
\thanks{This work was supported by 
JSPS KAKENHI Grant-in-Aid for Young Scientists Grant Number JP26K17382, 
the Okawa Publication Prize Grant Number 25-03 (to A. Nakai-Kasai), 
and JSPS KAKENHI Grant-in-Aid for Scientific Research(A) Grant Number JP25H01111 (to T. Wadayama).}% <-this % stops a space
}

\markboth{Journal of \LaTeX\ Class Files,~Vol.~14, No.~8, August~2021}%
{Shell \MakeLowercase{\textit{et al.}}: A Sample Article Using IEEEtran.cls for IEEE Journals}


\maketitle

\begin{abstract}
  % Federated learning (FL) is a distributed machine learning method 
  % that repeatedly updates the model on multiple clients and aggregates the models on a central server 
  % via wireless communication.
  % The heterogeneity of device states, statistical properties of data, and communication environments of clients exists in many FL applications 
  % and causes the degradation of inference accuracy of the learned model.
  % Therefore, it is necessary to appropriately adjust the aggregation weights used at the server for each client.
  % To address the problem of additional learning of aggregation weights using training data at each iteration of FL,
  % we propose a method that updates the aggregation weights online using hypergradients.
  Federated learning using mobile and Internet of Things devices requires not only the ability to handle heterogeneity of clients' data distributions 
  but also high adaptability to varying communication environments. 
  We propose FedHAW (Federated Learning with Hypergradient-based update of Aggregation Weights) 
  that implements online updates of aggregation weights.
  FedHAW updates the aggregation weights by using hypergradient, the gradient of the objective function with respect to the weights, 
  which can be calculated with low computational overhead.
  % a online update method of the FL training process by using gradient with respect to the aggregation weights.
  % The proposed method has both high generalization performance of the recently proposed FL method  
  % and high robustness to the learning environmental changes.
  Simulation results show that the proposed method possesses high generalization performance in heterogeneous environments
  and high robustness to communication errors.
\end{abstract}


\section{Introduction}
\IEEEPARstart{I}{ncreasing}
number of mobile and IoT (Internet of Things) devices encourages the provision of applications utilizing personal data stored on them.
FL (Federated Learning) \cite{McMahan} is a distributed machine learning method for such settings that achieves privacy protection and data processing load reduction:
each client updates its model on local data, the server aggregates the clients' models---primarily over wireless links---by weighted averaging, and the procedure repeats for multiple rounds without sharing raw personal data between the clients and the server.
% Federated Learning (FL) with Federated Averaging (FedAvg) \cite{McMahan} was proposed by Google.
% \begin{equation}
%   \bm{w}^{(t+1)}=\sum_{k=1}^K \theta_k\bm{w}_k^{(t)}, \ t=0,\ldots, T-1,
% \end{equation}
% where $\theta_k\geq0$ is the aggregation weight for client $k$ and $T$ is the number of rounds of FL training.

A serious problem in FL is heterogeneity of clients \cite{Wu}.
Clients have different computing capabilities and network environments.
The data distributions are non-IID (independent and identically distributed).
The heterogeneity causes bias in the learning process and degradation of the inference accuracy of the model \cite{Gafni}.
Especially in terms of IoT devices, the communication environment is often unstable due to its limited power and communication capability, 
and the erroneous communication link influences the FL training process \cite{error}.

To address this performance degradation due to the heterogeneity of clients,
FL algorithms that are robust to the learning environmental changes have been proposed.
The majority of such methods are based on the idea 
that the amount of each client's model parameter update in a round can be used to adjust each client's learning process.
Such amount is incorporated into the client update process \cite{fedprox,feddyn} 
or the central server's aggregation process \cite{fedadp,fedhyper}, 
and allows online update of the learning process according to the current environment.

FedLAW (Federated Learning with Learnable Aggregation Weights) \cite{FedLAW} is the method based on another perspective 
that achieves robustness to the learning environmental changes by learning of the FL training process itself.
% The method introduces the concept of weight decay in neural networks to the FL training process 
% and improves the generalization performance.
% In FedLAW, the aggregation weights $\theta_k$ are divided into relative weights $\bm{\lambda}=[\lambda_1,\ldots,\lambda_K]$ and its $\ell_1$ norm $\gamma=\|\bm{\lambda}\|_1$ as follows:
% \begin{equation}
%   \bm{w}^{(t+1)}=\gamma\left(\sum_{k}\lambda_k\bm{w}_k^{(t))}\right), \ \lambda_k\geq0, \ \gamma>0
%   \label{eq:fedlaw}
% \end{equation}
The update rules of FedLAW correspond to learning the aggregation weights in each round of FL training using pre-prepared data.
However, this overhead can be significant relative to total FL training time and may induce unexpected changes in communication or device conditions during training.
% The learning time overhead can cause unexpected communication environmental and device environmental changes.
% Furthermore, data specially for the aggregation weight learning must be prepared that is suitable for the purpose of the main FL process.
Another drawback is the need to prepare data specifically for the aggregate weight learning.
It is often difficult to prepare data that reflect the true data distribution.
% by dividing the personal data into learning data, 
% it is possible to affect the inference accuracy of the main FL process.

In this letter, 
we improve the robustness of FedLAW to the learning environmental changes 
by enabling online update of the aggregation weights.
We propose FedHAW (Federated Learning with Hypergradient-based update of Aggregation Weights), 
a method that updates the aggregation weights online using hypergradient descent \cite{Baydin}.
The hypergradient descent is an online learning rate optimization method for gradient descent 
using the gradient of the objective function with respect to the learning rate,
% This method has high robustness to the initial value choices of learning rate  
% and high tracking ability to the environmental changes.
which yields high tracking ability to the environmental changes.
% This idea can be extended to other hyperparameters and algorithms instead of the learning rate of gradient descent 
We apply this online update idea to the aggregation weights of FedLAW 
to reduce learning time overhead and improve adaptability to the current environment.
The main contributions of this letter are as follows:
\begin{itemize}
  \item We propose FedHAW that allows the hypergradient-based online update of the aggregation weights in each round of FL training process, 
  % by applying the hypergradient descent to the aggregation weights of FL algorithms.
  which enables adaptively adjusting the learning process to the heterogeneity of data distributions and the current communication environment with low computational overhead.
  FedHAW does not require additional training data.
  \item FedHAW can be combined with any client update process of FL algorithms.
  \item Simulation results show that FedHAW possesses both high generalization performance and online capability.
  % of FedLAW in heterogeneous environments
  % and high robustness to the communication erroneous.
\end{itemize}

\section{Preliminaries}
% \subsection{Notation}
% In the remainder of this paper, we use the following notations:
% The superscript $(\cdot)^{\mathrm{T}}$ denotes a transpose operation.
% The zero vector and the vector whose elements are all $1$ are represented by $\bm{0}$ and $\bm{1}$, respectively.
% The $\ell_1$ and Euclidean ($\ell_2$) norms are denoted by $\|\cdot\|_1$ and $\|\cdot\|_2$, respectively.
% , and $\ell_1$ norm is denoted by $\|\cdot\|_1$.
% The complex circularly symmetric Gaussian distribution $\mathcal{CN}(\bm{0},\bm{\Sigma})$ 
% has a mean vector $\bm{0}$ and a covariance matrix $\bm{\Sigma}$.
% The expectation and trace operators are $\mathbb{E}[\cdot]$ and $\mathrm{Tr}[\cdot]$, respectively.
% The expectation operator is $\mathbb{E}[\cdot]$.
% The diagonal matrix is given by $\mathrm{diag}[\ldots]$ with the diagonal elements shown in square brackets.
% The matrix exponential $\exp(\bm{A})$ for a matrix $\bm{A}$ is defined by 
% $\exp(\bm{A}):=\sum_{k=0}^\infty \frac{1}{k!}\bm{A}^k$.

\subsection{Federated Learning}
Let us consider a system comprising $K$ clients and a single central server. 
Each client has its own local dataset.
The number of local data samples for client $k \ (k=1,\ldots,K)$ is $N_k$, and the total number is $N=\sum_{k=1}^K N_k$.
% The communication network has star topology where the central server acts as the hub.
This letter focuses on generalization context of FL 
where the clients and the central server share the same objective 
of reducing the generalization error.
% of finding a set of learning model parameters $\bm{w}_\mathrm{opt}\in\mathbb{R}^d$ that minimizes a global objective function. 
% $f:\mathbb{R}^d\to\mathbb{R}$ 
% without sharing the local data, that is, 
% $\bm{w}_\mathrm{opt} = \mathrm{arg}\min_{\bm{w}\in\mathbb{R}^d} f(\bm{w})$.
% A typical objective function has the following finite-sum form: 
% \begin{align}
%   f(\bm{w}) = \sum_{k=1}^K \theta_k f_k(\bm{w}),
%   \label{eq:objective}
% \end{align}
% where $\theta_k\in\mathbb{R}$ is a nonnegative weight for client $k$ that satisfies $\sum_{k=1}^K \theta_k=1$, 
% and $f_k:\mathbb{R}^d\to\mathbb{R}$ is the client $k$'s local objective function 
% composed of loss function values.

In FL algorithms, 
the clients and the server iterate the following two procedures for $T$ rounds, 
(1) \emph{client update}: 
each client performs local updates of its model using its own local data, 
and (2) \emph{aggregation}: 
the server aggregates the clients' models obtained by the client update.
% FedAvg iterates client selection, 
% parallel client updates of the model parameters based on vanilla stochastic gradient descent (SGD) at the selected clients, 
% and model aggregation at the server via weighted averaging of the model parameters, 
% to collaboratively find the optimal model parameter.
% 
Let $\ell(\bm{B}_k;\bm{w})$ be client $k$'s loss on the model parameter $\bm{w}$ and 
its minibatch $\bm{B}_k$ from the local dataset, 
and $g(\nabla \ell(\bm{B}_k;\bm{w}))$ be a function of its gradient corresponding to SGD and Adam, etc.
The client update procedure at each batch is given by 
\begin{equation}
  \bm{w}_k^{(t)} \leftarrow \bm{w}_k^{(t)}-\eta g(\nabla \ell(\bm{B}_k;\bm{w}_k^{(t)})), 
  \label{eq:clientupdate}
\end{equation}
where $\eta$ is the learning rate.
After the client updates for pre-set $E$ epochs, 
in most of the FL algorithms, 
the server aggregates the local models using 
\begin{equation}
  \bm{w}^{(t+1)} = \sum_{k=1}^K \theta_k\bm{w}_k^{(t)}, \ t=0,\ldots, T-1,
\end{equation}
where $\theta_k$ is the nonnegative aggregation weight for client $k$ and the sum is set to $\sum_{k=1}^K \theta_k=1$.
The weight setting depends on the algorithm.
FedAvg \cite{McMahan} is the best-known federated learning algorithm 
and its weight is given by $\theta_k=N_k/N$.

% The procedure of FedAvg and the local client update based on vanilla stochastic gradient descent (SGD) are summarized in Algorithm~\ref{alg:fedavg}, 
% where $T$ is the number of iteration rounds, %$B$ is the size of the local minibatch, 
% $E$ is the number of local epochs, and $\eta$ is the learning rate.
% Note that the detailed processes for client selection are omitted 
% because they are beyond the focus of this paper.

% The weight 
% \begin{align}
%   \theta_k^\mathrm{Avg} = \frac{N_k}{N}  
%   \label{eq:thetafedavg}
% \end{align} 
% is used for the aggregation step at the server (line 5 of Algorithm~\ref{alg:fedavg}).
% % \begin{equation}
% %   \bm{w}^{(t+1)} = \sum_{k=1}^K \frac{N_k}{N} \bm{w}_k^{(t)} = \sum_{k=1}^K \theta_k^{\mathrm{Avg}} \bm{w}_k^{(t)}.
% % \end{equation}
% This weighting rule can be regarded as reflecting the concept that 
% a client with more data contributes more to learning 
% if the data distributions are IID and the clients are in a homogeneous environment.

\subsection{FedLAW}
FedLAW \cite{FedLAW} is the aggregation weight control method 
that divides the weights $\theta_k$ into relative weights $\bm{\lambda}=[\lambda_1, \ldots,\lambda_K]^\mathrm{T}$ and its $\ell_1$ norm $\gamma=\|\bm{\lambda}\|_1$.
The idea is on the basis of the facts 
that weights where the sum (i.e., $\ell_1$ norm) is smaller than 1 act as regularization for global model parameters and can improve generalization performance,
and that properly controlling relative weights by a certain round is crucial for performance improvement.
The parameters $\bm{\lambda}$ and $\gamma$ are optimized on the central server by using pre-prepared proxy dataset.

The same client update process as other FL methods is performed and then 
the aggregation process of FedLAW is given by 
\begin{equation}
  \bm{w}^{(t+1)} = \gamma^\ast\sum_{k=1}^K \lambda_k^\ast\bm{w}_k^{(t)}, \ t=0,\ldots, T-1,
\end{equation}
where $\gamma^\ast$ and $\bm{\lambda}^\ast=[\lambda_1^\ast,\ldots,\lambda_K^\ast]^\mathrm{T}$ are optimized in each round using proxy dataset 
aiming at 
\begin{align}
  \gamma^\ast, \bm{\lambda}^\ast &= \mathrm{arg}\min_{\gamma,\bm{\lambda}}\mathcal{L}_{\mathrm{proxy}}\left(\gamma\sum_{k=1}^K\lambda_k \bm{w}_k^{(t)}\right), \nonumber \\
& \mathrm{s.t.} \ \gamma>0, \lambda_k\geq0, \|\bm{\lambda}\|_1=1,
\end{align}
where $\mathcal{L}_{\mathrm{proxy}}(\bm{w})$ is the loss calculated by the same criterion as that of FL training 
using the proxy dataset and model parameter $\bm{w}$.

\subsection{Erroneous Communication Model}
\label{sec:error}
We consider in this letter the FL framework with erroneous communication and reuse of the current model.
This is analogous to the erroneous communication and reuse of old local updates considered in \cite{error}.
In this case, the aggregation process is given by 
\begin{equation}
  \bm{w}^{(t+1)}=\sum_{k\in\mathcal{R}{(t)}}\theta_k\bm{w}_k^{(t)} + \sum_{k\in\mathcal{N}{(t)}}\theta_k\bm{w}^{(t)},
  \label{eq:errormodel}
\end{equation}
where $\mathcal{R}{(t)}$ and $\mathcal{N}{(t)}$ are the client sets whose updated models can or cannot be received in round $t$.
The only difference from the model in \cite{error} is 
that the latest model parameter $\bm{w}^{(t)}$, instead of the last updates $\bm{w}_k^{(t-1)}$, 
is used as an alternative for clients 
whose updates are not received.
The model \eqref{eq:errormodel} can cover cases 
where a client's model cannot be received for multiple consecutive rounds.

We assume as the model in \cite{error} that the downlink is error-free 
because the central server, in many cases, is equipped with a high-performance communication module that can transmit signals at high power.

\subsection{Hypergradient Descent}
Consider the gradient descent method in a context unrelated to FL, which is given by
$\bm{x}^{(t+1)}=\bm{x}^{(t)}-\alpha\nabla f(\bm{x}^{(t)}), \ (t=0,1,\ldots,T-1)$,
where $\alpha>0$ is the learning rate.
Hypergradient descent \cite{Baydin} aims at optimizing the learning rate $\alpha$ in an online manner. 
The method introduces the iteration-dependent learning rate $\alpha_t$ 
and updates it to $\alpha_{t+1}$ by using the gradient with respect to itself of the objective function value at the next iteration, 
$\partial f(\bm{x}^{(t+1)})/\partial\alpha_t$, named hypergradient.
% it is assumed that the update rule for $\bm{x}^{(t+1)}$ can be used to optimize $\alpha_{t+1}$. 
The hypergradient can be calculated by using the chain rule:
\begin{equation}
	\frac{\partial f(\bm{x}^{(t+1)})}{\partial\alpha_t}=\frac{\partial f(\bm{x}^{(t+1)})}{\partial \bm{x}^{(t+1)}} \frac{\partial \bm{x}^{(t+1)}}{\partial\alpha_t},
  % =-\nabla f(\bm{x}^{(t+1)})^\mathrm{T}\nabla f(\bm{x}^{(t)}).
\end{equation}
and $\partial \bm{x}^{(t+1)}/\partial\alpha_t=-\nabla f(\bm{x}^{(t)})$.
The gradient descent with the hypergradient descent is given by:
\begin{equation}
	\bm{x}^{(t+1)}=\bm{x}^{(t)}-\alpha_t\nabla f(\bm{x}^{(t)}),
\end{equation}
% and the learning rate is updated by using the hypergradient:
\begin{equation}
	\alpha_{t+1}=\alpha_{t}-\beta\frac{\partial f(\bm{x}^{(t+1)})}{\partial\alpha_t}=\alpha_{t}+\beta\nabla f(\bm{x}^{(t+1)})^\mathrm{T}\nabla f(\bm{x}^{(t)}),
\end{equation}
where $\beta>0$ is the meta learning rate.

\section{Proposed Method: FedHAW}
\label{sec:fedhaw}
\subsection{Hypergradient-based Update of Aggregation Weights}
\label{sec:fedhaw_update}
The concept of hypergradient descent \cite{Baydin} that updates hyperparameters in an algorithm in an online manner can be extended to other hyperparameters and algorithms, 
instead of the learning rate of gradient descent as in the hypergradient descent.
We propose a method that applies the concept of hypergradient descent to the coefficients $\gamma,\bm{\lambda}$ in FedLAW.
In other words, the round-dependent coefficients $\gamma^{(t)},\bm{\lambda}^{(t)}$ are updated by using the hypergradients, respectively, 
i.e., $\partial f(\bm{w}^{(t)})/\partial \gamma^{(t)}$ and $\partial f(\bm{w}^{(t)})/\partial \lambda_k^{(t)}$.
The coefficients can be updated online during the execution of FL training process 
and this makes it possible to adaptively adjust the learning process to the current environment.

\begin{figure}[t]
  \centering
      \includegraphics[width=\columnwidth]{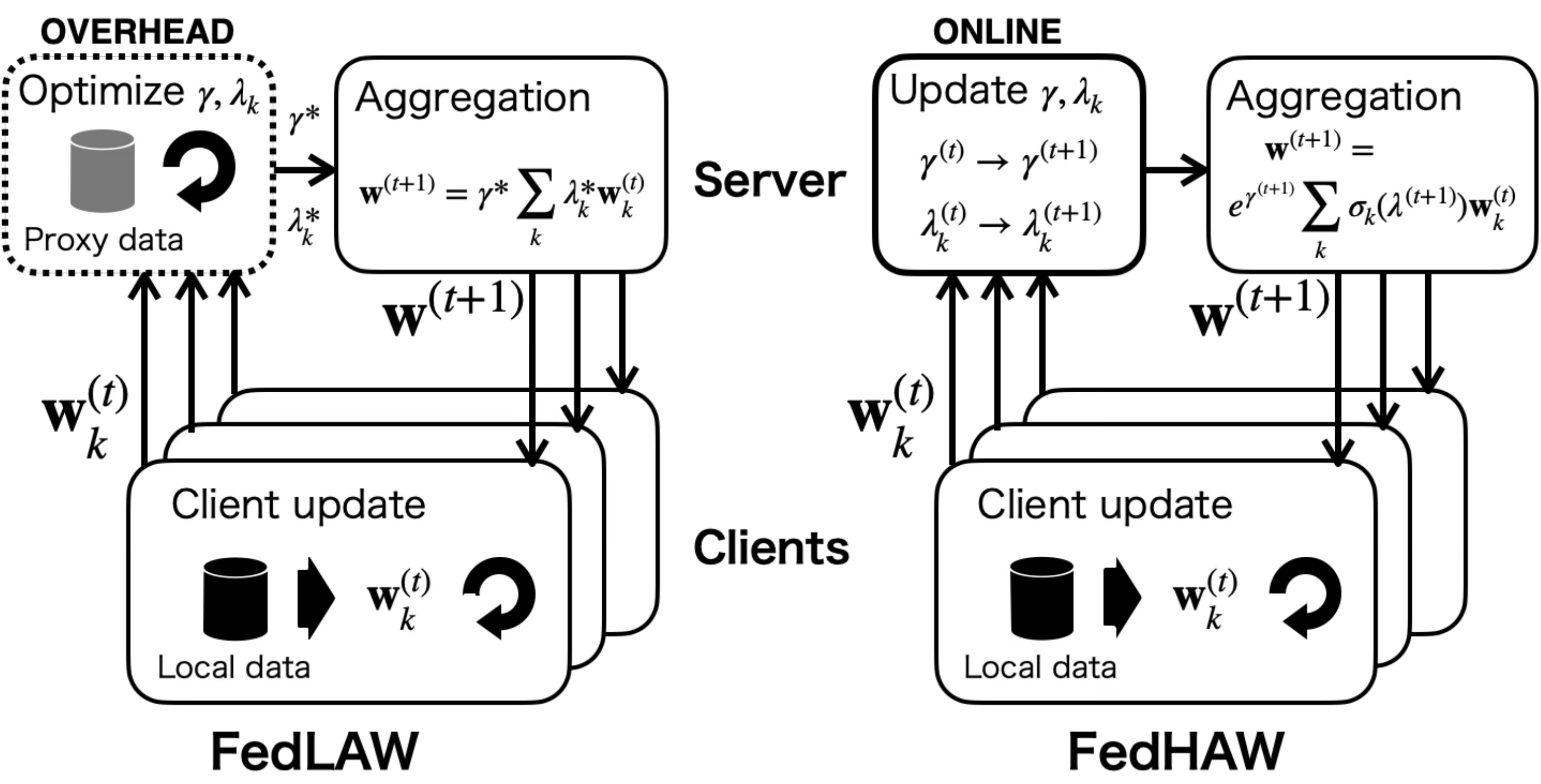}
    %  \vspace{2.0cm}
      \caption{Algorithm flows of FedLAW and the proposed FedHAW.}
      \label{fig:fedlawfedhaw}
\end{figure}
The algorithm of FedHAW is summarized in Algorithm~\ref{alg:fedhaw} 
and the difference from FedLAW is illustrated in Fig.~\ref{fig:fedlawfedhaw}.
Each coefficient is updated once in each round of FL training process.
The detailed calculation of the hypergradients is summarized in the following subsection.
\begin{algorithm}[tb]
  \caption{FedHAW}
  \label{alg:fedhaw}
  \begin{algorithmic}[1]
    \STATE Server: initialize model parameter $\bm{w}^{(0)}$, coefficients $\gamma^{(0)}$, $\bm{\lambda}^{(0)}$
    \FOR{$t=0,\ldots,T-1$}
      % \STATE Server: select $K$ clients and distribute $\bm{w}^{(t)}$
      \STATE Server: distribute $\bm{w}_k^{(t)}=\bm{w}^{(t)}$ to each client $k$
      \FOR{$e=0,\ldots,E-1$}
      	\STATE Each client $k$: $\bm{w}_k^{(t)} \leftarrow \bm{w}_k^{(t)}-\eta g(\nabla \ell(\bm{B}_k;\bm{w}_k^{(t)}))$
      \ENDFOR
      % \STATE Hypergradient-based update:
      % \STATE Same as FedLAW l3--l7
      \IF{$t>0$}
        \STATE Server: $\bm{d}^{(t)}=\bm{w}^{(t)}-\sum_{j}\sigma_j(\bm{\lambda^{(t)}})\bm{w}_j^{(t)}$
        \STATE Server: $\gamma^{(t+1)}=\gamma^{(t)}-\eta_\gamma \frac{e^{\gamma^{(t)}}}{\eta}\bm{d}^{(t)\mathrm{T}}\bm{w}^{(t)}$
        \STATE Server: \\$\lambda_k^{(t+1)}\!=\!\lambda_k^{(t)}\!- \frac{\eta_\lambda e^{2\gamma^{(t)}} \sigma_k(\bm{\lambda^{(t)}})(1-\sigma_k(\bm{\lambda^{(t)}}))}{\eta}\!\bm{d}^{(t)\mathrm{T}}\!\bm{w}_k^{(t-1)}$
      \ENDIF
      \STATE Server: $\bm{w}^{(t+1)}=e^{\gamma^{(t+1)}}\left(\sum_{k}\sigma_k(\bm{\lambda}^{(t+1)})\bm{w}_k^{(t)}\right)$
    \ENDFOR
    \RETURN $\bm{w}^{(T)}$
  \end{algorithmic}
\end{algorithm}

\subsection{Hypergradient Calculations}
\label{sec:fedhaw_calc}
First, we consider the aggregation formula incorporating the round-dependent constraints on $\gamma^{(t+1)}$ and $\bm{\lambda}^{(t+1)}$,
which is given by
\begin{equation}
  \bm{w}^{(t+1)}=e^{\gamma^{(t+1)}}\sum_{k=1}^K \sigma_k(\bm{\lambda}^{(t+1)})\bm{w}_k^{(t)}, 
\end{equation}
where $\sigma_k(\bm{\lambda}^{(t+1)})$ is the softmax function, $\sigma_k(\bm{\lambda}^{(t+1)})={e^{\lambda_k^{(t+1)}}}/{\sum_{j=1}^K e^{\lambda_j^{(t+1)}}}$, 
and $\bm{\lambda}^{(t+1)}=[\lambda_1^{(t+1)},\ldots,\lambda_K^{(t+1)}]^{\mathrm{T}}$.
Note that the constraint of $\lambda_k^{(t)}\geq0$ is relaxed to $\lambda_k^{(t)}>0$ for the sake of simplicity.
For the erroneous communication model \eqref{eq:errormodel}, $\lambda_k^{(t+1)}$ is prepared for $k\in\mathcal{R}{(t+1)}$ and $k\in\mathcal{N}{(t+1)}$.

The proposed FedHAW introduces, in each round of FL training process, the following update rules for $\gamma$ and $\lambda_k$:
\begin{equation}
	\gamma^{(t+1)}=\gamma^{(t)}-\eta_\gamma\frac{\partial f(\bm{w}^{(t)})}{\partial \gamma^{(t)}},
  \label{eq:gammaupdate}
\end{equation}
\begin{equation}
	\lambda_k^{(t+1)}=\lambda_k^{(t)}-\eta_\lambda\frac{\partial f(\bm{w}^{(t)})}{\partial \lambda_k^{(t)}},
  \label{eq:lambdaupdate}
\end{equation}
% をAlgorithm\ref{alg:fedhaw}にまとめる．
where $\eta_\gamma,\eta_\lambda>0$ are the meta learning rates for $\gamma$ and $\lambda_k$s, respectively.
The hypergradients with respect to $\gamma^{(t)}$ and $\lambda_k^{(t)}$ are given by
\begin{equation}
  \frac{\partial f(\bm{w}^{(t)})}{\partial \gamma^{(t)}}=\frac{\partial f(\bm{w}^{(t)})}{\partial \bm{w}^{(t)}}\frac{\partial \bm{w}^{(t)}}{\partial \gamma^{(t)}},
  \frac{\partial f(\bm{w}^{(t)})}{\partial \lambda_k^{(t)}}=\frac{\partial f(\bm{w}^{(t)})}{\partial \bm{w}^{(t)}}\frac{\partial \bm{w}^{(t)}}{\partial \lambda_k^{(t)}},
  \label{eq:hypergradients}
\end{equation}
respectively.

The former part $\partial f(\bm{w}^{(t)})/\partial \bm{w}^{(t)}$ in \eqref{eq:hypergradients} is the gradient of the objective function with respect to $\bm{w}^{(t)}$.
Since a full calculation of this gradient is computationally undesirable, we use an approximate calculation.
The aggregation process in round $t$ can be expanded as 
\begin{align}
  \bm{w}^{(t+1)}
  % &\!=\!e^{\gamma^{(t+1)}}\!\left(\!\bm{w}^{(t)}\!-\!\eta\!\sum_{k}\!\sigma_k(\bm{\lambda}^{(t+1)})\!\sum_{b}\! g(\nabla \ell(\bm{B}_k^{[b]};\bm{w}_k^{(t)[b]}))\right) \nonumber\\
  &= \bm{w}^{(t)}-\eta e^{\gamma^{(t+1)}}\sum_{k}\sigma_k(\bm{\lambda}^{(t+1)})\left(\frac{\bm{w}^{(t)}-\bm{w}_k^{(t)}}{\eta}\right) \nonumber\\
  & \quad -(1-e^{\gamma^{(t+1)}})\bm{w}^{(t)},
  \label{eq:aggregationprocess}
\end{align}
% where the sum over $b$ means the iterations of the client update \eqref{eq:clientupdate} 
% and where the equation $\sum_{b} g(\nabla \ell(\bm{B}_k^{[b]};\bm{w}_k^{(t)[b]}))=-(\bm{w}_k^{(t)}-\bm{w}^{(t)})/\eta$ holds by using \eqref{eq:clientupdate}.
where the amount of local model updates in the client update can be expressed as $-(\bm{w}_k^{(t)}-\bm{w}^{(t)})/\eta$ by using \eqref{eq:clientupdate}.
The third term can be viewed as weight decay term for the model parameter $\bm{w}^{(t-1)}$, which is discussed in \cite{FedLAW}.
We consider the portion of the second term excluding the learning rate $\eta$ to be the gradient of the objective function with respect to $\bm{w}^{(t)}$.
Therefore, the approximate gradient is given by
\begin{equation}
  \left(\frac{\partial f(\bm{w}^{(t)})}{\partial \bm{w}^{(t)}}\right)^\mathrm{T}\approx e^{\gamma^{(t+1)}}\frac{\bm{w}^{(t)}-\sum_{k}\sigma_k(\bm{\lambda}^{(t+1)})\bm{w}_k^{(t)}}{\eta}.
  \label{eq:approxgradient}
\end{equation}

The latter parts $\partial \bm{w}^{(t)}/\partial \gamma^{(t)},\partial \bm{w}^{(t)}/\partial \lambda_k^{(t)}$ in \eqref{eq:hypergradients} are the gradients of the aggregated model $\bm{w}^{(t)}$ in previous round, i.e., $\bm{w}^{(t)}=e^{\gamma^{(t)}}\sum_{k=1}^K \sigma_k(\bm{\lambda}^{(t)})\bm{w}_k^{(t-1)}$.
These can be calculated directly as follows:
\begin{equation}
  \frac{\partial \bm{w}^{(t)}}{\partial \gamma^{(t)}}=e^{\gamma^{(t)}}\sum_{k=1}^K \sigma_k(\bm{\lambda}^{(t)})\bm{w}_k^{(t-1)}=\bm{w}^{(t)},
  \label{eq:gammat}
\end{equation}
\begin{equation}
  \frac{\partial \bm{w}^{(t)}}{\partial \lambda_k^{(t)}}=e^{\gamma^{(t)}}\sigma_k(\bm{\lambda}^{(t)})(1-\sigma_k(\bm{\lambda}^{(t)}))\bm{w}_k^{(t-1)}.
  \label{eq:lambdat}
\end{equation}

% Using these results, the hypergradients of $\gamma$ and $\lambda_k$ can be calculated as follows:
% \begin{equation}
% \frac{\partial f(\bm{w}^{(t)})}{\partial \gamma}=\frac{e^\gamma}{\eta}(\bm{w}^{(t)}-\sum_{k}\sigma_k(\bm{\lambda})\bm{w}_k^{(t)})^\mathrm{T}\bm{w}^{(t)}
% \end{equation}
% \begin{equation}
% 	\frac{\partial f(\bm{w}^{(t)})}{\partial \lambda_k}=\frac{e^{2\gamma} \sigma_k(\bm{\lambda})(1-\sigma_k(\bm{\lambda}))}{\eta}(\bm{w}^{(t)}-\sum_{k}\sigma_k(\bm{\lambda})\bm{w}_k^{(t)})^\mathrm{T}\bm{w}_k^{(t-1)}
% \end{equation}
By substituting \eqref{eq:approxgradient}--\eqref{eq:lambdat} into \eqref{eq:hypergradients}, 
the hypergradients in \eqref{eq:gammaupdate} and \eqref{eq:lambdaupdate} can be calculated.
The proposed FedHAW can update the coefficients $\gamma,\bm{\lambda}$ online during the execution of FL training process,
using only the model parameters $\bm{w}^{(t)},\bm{w}_k^{(t)},\bm{w}_k^{(t-1)}$ already obtained,
and thus the aggregation weight adjustment to the current environment can be performed with low computational load.

The update rule of FedHAW does not depend on the client update process of FL algorithms 
since the update rule is only based on the local model differences $\bm{w}^{(t)}-\bm{w}_k^{(t)}$ in \eqref{eq:aggregationprocess} 
and it does not depend on the method used to update the local model.
In other words, FedHAW can be combined with any client update process of FL algorithms 
such as that adopted in \cite{fedprox}.

\section{Simulation Results}
\subsection{Experimental Setup}
In this section, we evaluate the image classification performance of the proposed FedHAW under heterogeneous environments and communication errors.
We prepared non-IID local datasets for $K=10$ clients from MNIST dataset, CIFAR-10 dataset, and Stanford Dogs dataset \cite{dogs1,dogs2}
by using the Dirichlet distribution \cite{mnist,Yurochkin}.
The concentration parameter $D_\alpha$ of the Dirichlet distribution was set to $D_\alpha\in\{1,0.1\}$. 
% where a smaller value promotes clients holding data for specific classes and enhances non-IIDness.
% We used cross-entropy loss for all datasets.
We used a 3-layer MLP for MNIST, ResNet-18 for CIFAR-10, 
and a frozen ImageNet-21k-pretrained ViT \cite{vit} + FC for Stanford Dogs.
Table~\ref{tab:exp_setup} summarizes the dataset-specific local settings.
\begin{table}[tb]
\caption{Dataset-specific local training settings.}
\label{tab:exp_setup}
\centering
% \scriptsize
\begin{tabular}{@{}l|ccccc@{}}
\hline\hline
Dataset & $T$ & Epochs & Batch & $\eta$ & Optimizer \\
\hline
MNIST & 200 & 1 & 64 & $10^{-3}$ & SGD \\
CIFAR-10 & 50 & 10 & 64 & $10^{-1}$ & SGD+WD \\
Dogs & 100 & 10 & 64 & $5.0\times10^{-5}$ & Adam \\
\hline\hline
\end{tabular}
\end{table}

The meta learning rates for FedHAW were set to $(\eta_\gamma,\eta_\lambda)=(10^{-3},10^{-2})$ for MNIST dataset, 
$(10^{-5},10^{-2})$ for CIFAR-10 dataset, and $(10^{-7},10^{-5})$ for Stanford Dogs dataset.
The initial parameters were set to $\gamma^{(0)}=0$ and $\lambda_k^{(0)}=N_k/N$.
% For each dataset, we partitioned the test split into two disjoint subsets: 
% a proxy subset of size \(10\times\) the number of classes (used only by FedLAW) 
% and the remaining subset for final evaluation of all methods.

We compared performance with conventional FL algorithms in terms of the server-side aggregation process, 
FedAvg \cite{McMahan}, FedAdp \cite{fedadp}, FedHyper-G \cite{fedhyper}, FedLWS \cite{fedlws}, and FedLAW \cite{FedLAW}.
With the exception of FedLAW, these methods are online methods that do not require separate training data or separate training in every round.
For FedLAW, we used a proxy dataset of size $10\times$ the number of classes from each dataset's test split, 
with $100$ proxy epochs and learning rate $=0.01$.
No sample in the proxy subset was used in the final test subset for all methods.
FedLWS is an extension of FedLAW that uses layer-dependent $\gamma$ and employs online learning for it.
% but no design guidelines regarding the weight $\lambda_k$ were provided (meaning it can be combined with any weight), 
% so 
In this experiment, we used the same weight as in FedAvg for FedLWS.
We report test accuracy on held-out balanced test sets for each dataset.
% Test accuracy was evaluated 
% for the model obtained through federated learning 
% using class-balanced test data not included in the training data.
The experimental codes and datasets are available in a GitHub repository\footnote{https://github.com/a-nakai-k/FedHAW}.

\subsection{Performance and Execution Time Evaluation Under Heterogeneous Environments}
We first evaluate the performance of the proposed method in the presence of data heterogeneity 
under error-free communications.
Table~\ref{tab:heterogeneous} shows the test accuracy of methods for $D_\alpha=1$ and $0.1$ in each dataset.
The bold and underline values indicate the best and second best results, respectively; the same applies to the tables that follow.
FedLAW optimizes the parameters $\bm{\lambda}$ and $\gamma$ in each round using pre-prepared data at the expense of the learning time overhead,
so that the performance was the best or second best in most cases.
The proposed FedHAW achieved the best performance among the online methods in most cases.
In some cases, the proposed FedHAW outperforms FedLAW, 
which is due to the fact that the proxy dataset used in FedLAW does not necessarily reflect the true data distribution.
The results indicate that FedHAW is robust to the data heterogeneity.
\begin{table}[!t]
\caption{Test accuracy (\%) under heterogeneous data distributions.}
\label{tab:heterogeneous}
\centering
\begin{tabular}{@{}l|cc|cc|cc@{}}
\hline\hline
Dataset & \multicolumn{2}{c}{MNIST} & \multicolumn{2}{c}{CIFAR-10} & \multicolumn{2}{c}{Dogs} \\
$D_\alpha$ & 1 & 0.1 & 1 & 0.1 & 1 & 0.1 \\
 \hline
FedAvg \cite{McMahan} & $90.10$ & $82.96$ & $75.64$ & $64.10$ & $\underline{88.85}$ & $83.26$ \\
FedAdp \cite{fedadp} & $90.09$ & $83.16$ & $74.53$ & $59.92$ & $88.68$ & $83.84$ \\
FedHyper \cite{fedhyper} & $90.28$ & $83.71$ & $76.09$ & $55.29$ & $87.86$ & $82.98$ \\
FedLWS \cite{fedlws} & $90.04$ & $82.95$ & $78.31$ & $63.58$ & $88.75$ & $83.16$ \\
FedHAW & $\mathbf{91.55}$ & $\mathbf{88.16}$ & $\underline{80.15}$ & $\underline{64.88}$ & $88.13$ & $\mathbf{86.76}$ \\
\hline
FedLAW \cite{FedLAW} & $\underline{91.01}$ & $\underline{85.75}$ & $\mathbf{83.01}$ & $\mathbf{70.30}$ & $\mathbf{88.95}$ & $\underline{85.90}$ \\
\hline\hline
\end{tabular}
\end{table}

In addition, we evaluate the execution time of the methods for the aggregation process at the server side.
The execution time was measured on Stanford Dogs dataset with $D_\alpha=0.1$,
and on a machine with Intel(R) Xeon(R) w7-2495X CPU @ 2.5GHz, 128GB memory, and an NVIDIA RTX A4000 GPU.
Table~\ref{tab:executiontime} shows the average execution time of $T=100$ rounds.
FedLAW is the only method that requires optimization in each round 
so that it took $10^2$ to $10^4$ times longer than the other methods.
This substantiates the overhead requirement when running FedLAW.
On the other hand, 
the execution time of the proposed FedHAW is only four times that of FedAvg, 
the simplest aggregation method.
% indicating that aggregation of FedHAW can be performed in online manner.
The results in Tables~\ref{tab:heterogeneous} and~\ref{tab:executiontime} suggest that 
FedHAW is a practical solution that simultaneously achieves high performance and online capabilities.
\begin{table}[!t]
\caption{Execution time (s) for the server side aggregation process.}
\label{tab:executiontime}
\centering
\begin{tabular}{@{}c|c|c|c|c||c@{}}
  \hline
FedAvg & FedAdp & FedHyper & FedLWS & FedHAW & FedLAW \\
 \hline
$0.00014$ & $0.01090$ & $0.00026$ & $0.01925$ & $0.00056$ & $2.31493$ \\
  \hline
\end{tabular}
\end{table}

\subsection{Performance Evaluation Under Communication Errors}
\label{sec:exp_communicationerrors}
This subsection introduces the erroneous communication model of Sect. \ref{sec:error}.
% into FL training process and evaluates the performance.
We assumed that each client has its own communication error probability 
set from a uniform distribution with a maximum probability of $p_e$, 
and that each client fails to transmit its local model update to the server in accordance with these probabilities.
In such cases, model aggregation at the server follows \eqref{eq:errormodel}.
We set $p_e=0.2, 0.8$, and $D_\alpha=0.1$ for each dataset 
and evaluated the test accuracy.
The meta learning rates for FedHAW for MNIST were set to $(\eta_\gamma,\eta_\lambda)=(10^{-4},10^{-2})$ 
but the other parameter settings were kept the same as in the previous section.
Note that FedAdp is removed from the comparison 
because it requires local model updates in each round and 
does not support the model \eqref{eq:errormodel}.

\begin{figure}[tb]
\centering
\includegraphics[width=\columnwidth]{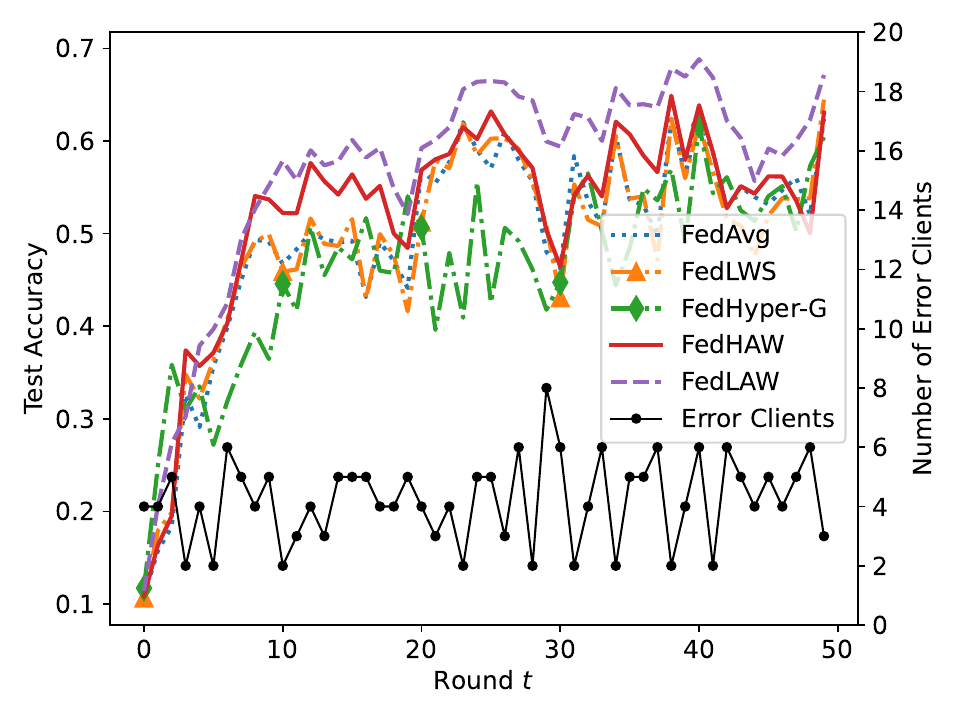}
\caption{Per-round accuracy and the number of clients with communication errors for CIFAR-10 with $p_e=0.8$.}
\label{fig:cifar10pe08}
\end{figure}

Table~\ref{tab:communicationerrors} shows the test accuracy of methods under communication errors.
As with the results in the previous subsection, the proposed FedHAW achieved the best performance among the online methods in most cases.
Figure~\ref{fig:cifar10pe08} shows per-round accuracy and the number of clients with communication errors
for CIFAR-10 with $p_e=0.8$.
The figure shows that the proposed FedHAW can maintain higher accuracy than the other methods 
even when the number of clients with communication errors is high, 
especially in the early rounds.
Particularly for IoT devices, 
which require ultra-fast response even under unstable network conditions, 
applying FedLAW that involves training on a proxy dataset every round is not practical. 
Therefore, we conclude that the proposed FedHAW could serve as a practical solution for such applications.
\begin{table}[tb]
\caption{Test accuracy (\%) under communication errors.}
\label{tab:communicationerrors}
\centering
\begin{tabular}{@{}l|cc|cc|cc@{}}
\hline\hline
Dataset & \multicolumn{2}{c}{MNIST} & \multicolumn{2}{c}{CIFAR-10} & \multicolumn{2}{c}{Dogs} \\
$p_e$ & 0.2 & 0.8 & 0.2 & 0.8 & 0.2 & 0.8 \\
 \hline
FedAvg \cite{McMahan} & $80.53$ & $73.41$ & $61.30$ & $62.39$ & $83.60$ & $82.33$ \\
FedHyper \cite{fedhyper} & $81.13$ & $75.17$ & $58.55$ & $60.40$ & $82.02$ & $83.84$ \\
FedLWS \cite{fedlws} & $80.52$ & $73.39$ & $63.44$ & $\underline{64.46}$ & $83.26$ & $81.58$ \\
FedHAW & $\mathbf{86.70}$ & $\mathbf{83.72}$ & $\underline{67.47}$ & $63.07$ & $\mathbf{87.34}$ & $\underline{85.21}$ \\
\hline
FedLAW \cite{FedLAW} & $\underline{83.26}$ & $\underline{78.98}$ & $\mathbf{68.39}$ & $\mathbf{67.12}$ & $\underline{86.14}$ & $\mathbf{85.80}$ \\
\hline\hline
\end{tabular}
\end{table}
% \begin{table*}[tb]
% \caption{Test accuracy (\%) under communication errors.}
% \label{tab:communicationerrors}
% \centering
% \begin{tabular}{@{}l|ccc|ccc|ccc@{}}
% \hline\hline
% Dataset & \multicolumn{3}{c}{MNIST} & \multicolumn{3}{c}{CIFAR-10} & \multicolumn{3}{c}{Dogs} \\
% $p_e$ & 0.2 & 0.5 & 0.8 & 0.2 & 0.5 & 0.8 & 0.2 & 0.5 & 0.8 \\
%  \hline
% FedAvg \cite{McMahan} & $80.53$ & $80.51$ & $73.41$ & $61.30$ & $64.64$ & $62.39$ & $83.60$ & $82.95$ & $82.33$ \\
% FedHyper \cite{fedhyper} & $81.13$ & $82.70$ & $75.17$ & $58.55$ & $57.26$ & $60.40$ & $82.02$ & $84.53$ & $83.84$ \\
% FedLWS \cite{fedlws} & $80.52$ & $80.68$ & $73.39$ & $63.44$ & $64.92$ & $\underline{64.46}$ & $83.26$ & $82.54$ & $81.58$ \\
% FedHAW & $\mathbf{86.70}$ & $\underline{83.00}$ & $\mathbf{83.72}$ & $\underline{67.47}$ & $\underline{66.32}$ & $63.07$ & $\mathbf{87.34}$ & $\mathbf{87.14}$ & $\underline{85.21}$ \\
% \hline
% FedLAW \cite{FedLAW} & $\underline{83.26}$ & $\mathbf{83.53}$ & $\underline{78.98}$ & $\mathbf{68.39}$ & $\mathbf{67.58}$ & $\mathbf{67.12}$ & $\underline{86.14}$ & $\underline{85.56}$ & $\mathbf{85.80}$ \\
% \hline\hline
% \end{tabular}
% \end{table*}

\subsection{Combination with Other Client Update}
As discussed in the last paragraph in Sect. \ref{sec:fedhaw_calc}, 
the hypergradient-based update of aggregation weights in FedHAW can be combined with any client update process of FL algorithms.
We finally evaluate the performance of the proposed FedHAW when combined with another client update process, FedProx \cite{fedprox}.
FedProx is a client update process that adds a proximal term to the loss function to regularize the model parameters.
% To apply the client update of FedProx specifically to FedHAW, 
% all that is required is to replace the loss function used in the client update on line 5 of Algorithm 1 with the one from FedProx.
We set $p_e=0.5$ and $D_\alpha=0.1$ for each dataset, 
and the other parameter settings were the same as in Sect. \ref{sec:exp_communicationerrors}.

Table~\ref{tab:fedprox} summarizes the test accuracy of FedProx and FedProx+HAW in rounds 
$t_i=iT/5-1$ for $i=1,2,3,4,5$.
% $t=T/5-1, 2T/5-1, 3T/5-1, 4T/5-1, T-1$.
FedProx+HAW shows better performance than FedProx in most rounds.
The fact that FedProx+HAW achieved higher accuracy, particularly from the early rounds,
indicates that the adoption of the hypergradient-based update can effectively improve the accuracy of FedProx.
\begin{table}[tb]
\caption{Test accuracy (\%) in round $t$ of FedProx and FedProx+HAW.}
\label{tab:fedprox}
\centering
\begin{tabular}{@{}c|r|ccccc@{}}
\hline\hline
& Round $t_i$ & $t_1$ & $t_2$ & $t_3$ & $t_4$ & $t_5$ \\
\hline
MNIST & FedProx & $49.24$ & $65.87$ & $75.46$ & $76.74$ & $81.36$ \\
& +HAW & $\mathbf{74.30}$ & $\mathbf{83.28}$ & $\mathbf{85.00}$ & $\mathbf{82.56}$ & $\mathbf{83.69}$ \\
\hline
CIFAR-10 & FedProx & $53.56$ & $59.03$ & $52.92$ & $61.30$ & $\mathbf{65.20}$ \\
& +HAW & $\mathbf{60.44}$ & $\mathbf{62.67}$ & $\mathbf{59.40}$ & $\mathbf{62.93}$ & $64.47$ \\
\hline
Dogs & FedProx & $66.96$ & $78.01$ & $\mathbf{80.62}$ & $81.23$ & $83.40$ \\
& +HAW & $\mathbf{81.58}$ & $\mathbf{85.52}$ & $74.96$ & $\mathbf{86.00}$ & $\mathbf{86.72}$ \\
\hline\hline
\end{tabular}
\end{table}

\section{Conclusions}
We proposed FedHAW that updates the FedLAW-style aggregation coefficients $\gamma$ and $\bm{\lambda}$ in an online manner.
FedHAW requires no proxy dataset or per-round auxiliary training, and can be combined with any client-side optimizer.
Simulation results under data heterogeneity and communication errors showed that 
FedHAW has robustness to such environmental issues.
The proposed idea of the hypergradient-based update can be extended to any FL algorithm, 
as long as the coefficients are differentiable.

\vfill

\end{document}